\documentclass[11pt]{article}
\usepackage{graphicx} 
\usepackage[utf8]{inputenc}
\usepackage{amsmath,amssymb}
\usepackage[top=1in,bottom=1in,left=1in,right=1in]{geometry}
\usepackage{sfmath}
\usepackage[super]{natbib}

\usepackage{titlesec}
\usepackage{authblk}
\usepackage{url}

 \renewcommand{\Authfont}{\fontfamily{phv}\fontsize{11}{13}\fontfamily{phv}\selectfont}
 \renewcommand{\Affilfont}{\fontfamily{phv}\fontsize{10}{12}\selectfont}

\usepackage{setspace}
\usepackage{float}
 
\title{PATCH: a deep learning method to assess heterogeneity of
artistic practice in historical paintings}

\author[1,2]{Andrew Van Horn*}
\author[2,3]{Lauryn Smith}
\author[4]{Mahamad Salah Mahmoud}
\author[1]{Michael McMaster}
\author[2]{Clara G. Pinchbeck}
\author[1,5]{Ina T. Martin}
\author[1,5]{Andrew Lininger}
\author[6]{Anthony Ingrisano}
\author[7]{Adam Lowe}
\author[7]{Carlos Bayod}
\author[2]{Elizabeth S. Bolman}
\author[1]{Kenneth Singer}
\author[1]{Michael Hinczewski}

\affil[1]{Department of Physics, Case Western Reserve University, 2076 Adelbert Rd. Cleveland, OH,
USA}

\affil[2]{Department of Art History and Art, Case Western Reserve University, 11201 Euclid Ave., Cleveland, OH, USA}

\affil[3]{The Frick Pittsburgh Museums and Gardens, 7227 Reynolds St.
Pittsburgh, PA, USA}

\affil[4]{Department of Computer and Data Sciences, Case Western Reserve
University, 10900 Euclid Ave., Cleveland, OH, USA}

\affil[5]{MORE Center, Case Western Reserve University, 10900 Euclid Ave., Cleveland, OH, USA}

\affil[6]{Painting Department, Cleveland Institute of Art, 11610 Euclid Ave., Cleveland, OH,USA}

\affil[7]{Factum Foundation, Calle Albarracín 28, Madrid, Spain}

\date{}

\renewcommand{\figurename}{\fontfamily{phv}\fontsize{9}{12}\selectfont \textbf{Figure}}
\renewcommand{\thetable}{\textbf{\arabic{table}}}
\renewcommand{\thefigure}{\textbf{\arabic{figure}}}
\renewcommand{\tablename}{\fontfamily{phv}\fontsize{9}{12}\selectfont \textbf{Table}}

\begin{document}
\fontfamily{phv}\selectfont
\maketitle
\emergencystretch 3em

\titleformat*{\section}{\fontsize{11}{13}\bfseries\selectfont}


\textbf{In the Renaissance and Early Modern period, paintings were largely produced by master painters who directed workshops of apprentices and others who often contributed to the piece. Discerning who created these masterworks and how they did so is a central question in technical art history\cite{van1993search} and a nontrivial problem that machine learning can help solve by extending analysis to a microscopic scale\cite{ji2021discerning}. Analysis of workshop paintings presents a challenge, however, because information about the members of workshops and the processes by which artworks were created remains elusive\cite{bambach1999drawing,brooks2015andrea}. Thus, external examples are not available to train networks to recognize. Here we present a novel machine learning approach we call pairwise assignment training for classifying heterogeneity (PATCH) that is capable of identifying individual artistic practice regimes with no external training data. We apply this method to two historical paintings by the Spanish Renaissance master, El Greco, and our findings regarding one of the works potentially challenge previous studies that assert that a considerable portion of the painting was completed by workshop members after El Greco's death\cite{marias2013elgreco}. PATCH outperforms statistical and unsupervised machine learning methods in this complex pairwise comparison problem lacking “ground truth'' data, making it potentially useful across similar cases in the social and natural sciences, including image segmentation in remote sensing\cite{khanal2020remote}, urban development and design\cite{francis2023unsupervised}, and anomaly detection manufacturing contexts\cite{kharitonov2022comparative,liu2024deep}, among others.} 

\vspace{1em}

There are many ways to create a painting. The process can be
collaborative or solitary, and there is no end to the styles, tools,
materials and techniques that can be implemented. The evolution of what
we call artistic practice---the technical, stylistic, material, and even
physiological aspects of an artist's creative process---over time and its
variation between artists remains a key element of technical art history
scholarship. From the late Middle Ages through the Early Modern period, on the other hand, painting was typically done in a bustling
workshop\cite{alpers2005vexations}. A master painter would establish a workshop and assemble a group of apprentices, journeymen and specialist painters and even collaborate with associates to help fulfill commissions and
create smaller works to support the enterprise between larger contracts.

Workshop practice adds a second layer of complexity to artistic
practice. Master painters varied considerably in their managerial
styles. Some performed much of the work themselves, while others
dictated the composition but left many elements to the hands of workshop
members (e.g., van Dyck\cite{brown1983van}). Still others involved
apprentices in the conceptualization of a piece\cite{talvacchia2007raphael}.
Often, there was a hierarchy, with young apprentices typically assigned
more mundane tasks such as the mixing of paints and the construction of
canvases and more senior workshop members directed to paint portions of
larger works in the master's style\cite{bambach1999drawing}.

Because workshops varied so widely in their organization and management
yet produced works with cohesive styles, deeper understanding of
variation in artistic practice can shed light on workshop practice and
vice versa. Research into these key aspects of Renaissance and Early
Modern art production is hindered by a lack of surviving textual sources
regarding the structure, size, and membership of individual workshops as
well as their day-to-day operation and the variation in artistic
practices\cite{bambach1999drawing}. While scholars have been able to
reconstruct individual artists' workshops through a variety of methods,
our collective understanding of the inner workings of workshops is still
fragmentary\cite{brooks2015andrea}. Artistic practice varies widely,
encompassing materials, methods, styles, and individual contributions.
Understanding who interacted with individual paintings and how they did
so---discerning the artists at work in a painting, the interplay of their
individual ``hands,'' and the interaction of the artist's hand and eye
with paint and brush\cite{vanHoogstraten2012inleyding}---is an important open question
in art history\cite{van1993search}. 

Machine learning has recently emerged as a promising complement to
traditional art historical analyses in applications including
preservation, conservation, and the detection of forgeries\cite{hughes2010quantification,friedman2012authentication,kogou2020remote,assael2022restoring}, as well as in fighting illegal trafficking in
antiquities\cite{winterbottom2022deep}. Much work involving style is focused on
broad classification of works\cite{sandoval2019two,bigerelle2023fractal}, such as the use of
a neural network to arrange works of the Western canon in the correct
chronological order\cite{elgammal2018shape}. Recently published work by some
of the authors of the present study demonstrated the efficacy of ML for
attribution of paintings by analyzing high-resolution topographic images
of the surface texture of oil paintings by known authors (student
painters) in a controlled experimental setting. In a supervised machine
learning analysis, a convolutional neural network (CNN) was trained to
sort microscopic surface textures among artists, assigning 1
cm$^2$ patches of paintings to the correct authors with
$\sim$95\% accuracy\cite{ji2021discerning}. Thus, the method has
shown that microscopic analysis can complement traditional art
historical analyses of macroscopic features.

Unfortunately, supervised deep learning on topographic images cannot be used 
when there is no ground truth for the network to learn on. When the object of analysis is a painting produced by a Renaissance or Early Modern artist's workshop, the number of
artistic practice regimes is generally unknown, known examples of
the included artists' work may not exist, and knowledge of the painting
practices employed is incomplete. Unsupervised learning, wherein the machine analyzes individual objects (e.g., images or patches of an image) and creates classes based on the
statistical properties of those objects, would appear more appropriate. However, a naive unsupervised analysis still depends on hyperparameters that can be tuned to yield larger or smaller numbers of classes. Whether performed by the trained art historian or by a neural network, the problem of identifying regions of shared artistic practice is, in a word, nontrivial.

Here, we demonstrate a novel technique, to our knowledge, for ML-based image attribution, and use it to assign paintings to their respective artist(s). We then combine this method with network analysis to identify artistic practice regimes
(combinations of artists and materials) and create a measure of the
heterogeneity of artistic practice (HAP) within a given painting or set
of paintings. We apply this method to two important historical paintings
by El Greco, one considered to be entirely by the master
himself, and one previously thought to feature the work of members of
his workshop.

The method described here, Pairwise Assignment Training for Classifying Heterogeneity (PATCH), uses supervised learning toward an unsupervised end. Rather than train a network to recognize individual classes, we test whether the network is \emph{capable} of learning to distinguish between two objects and then use that information to build \emph{post-hoc} classes. The PATCH method relies on the inability of the network to correctly sort patches of paintings or regions of a painting painted by the same artist. Given patches of two paintings by different painters---a different-artist pair---the network will learn to assign those patches to the correct painting with a high degree of accuracy, as demonstrated in our previous publication. But if the network is given patches of two paintings by the same artist---a same-artist pair---it will not be able to learn to assign those patches to the correct painting; it will perform no better than a coin flip. We then create a network of same-artist pairs and use network analysis to construct
classes corresponding to artistic practice regimes. In this way, we
obviate the need for separate ground truth information for individual
artists and artistic practice regimes, enabling successful applications
in contexts such as workshop paintings where the ground truth is not
known with certainty.

\section*{Development and validation of the PATCH algorithm}

The PATCH method (Figure 1) consists of two phases. The first phase, pairwise
assignment training, identifies pairs of paintings or pairs of regions
of a single painting that are the ``same,'' that is, created by the same
artist under the same conditions. The second phase, community finding,
creates \emph{post hoc} classes based on these pairings.

\begin{figure}[t]
\includegraphics[width=\textwidth]{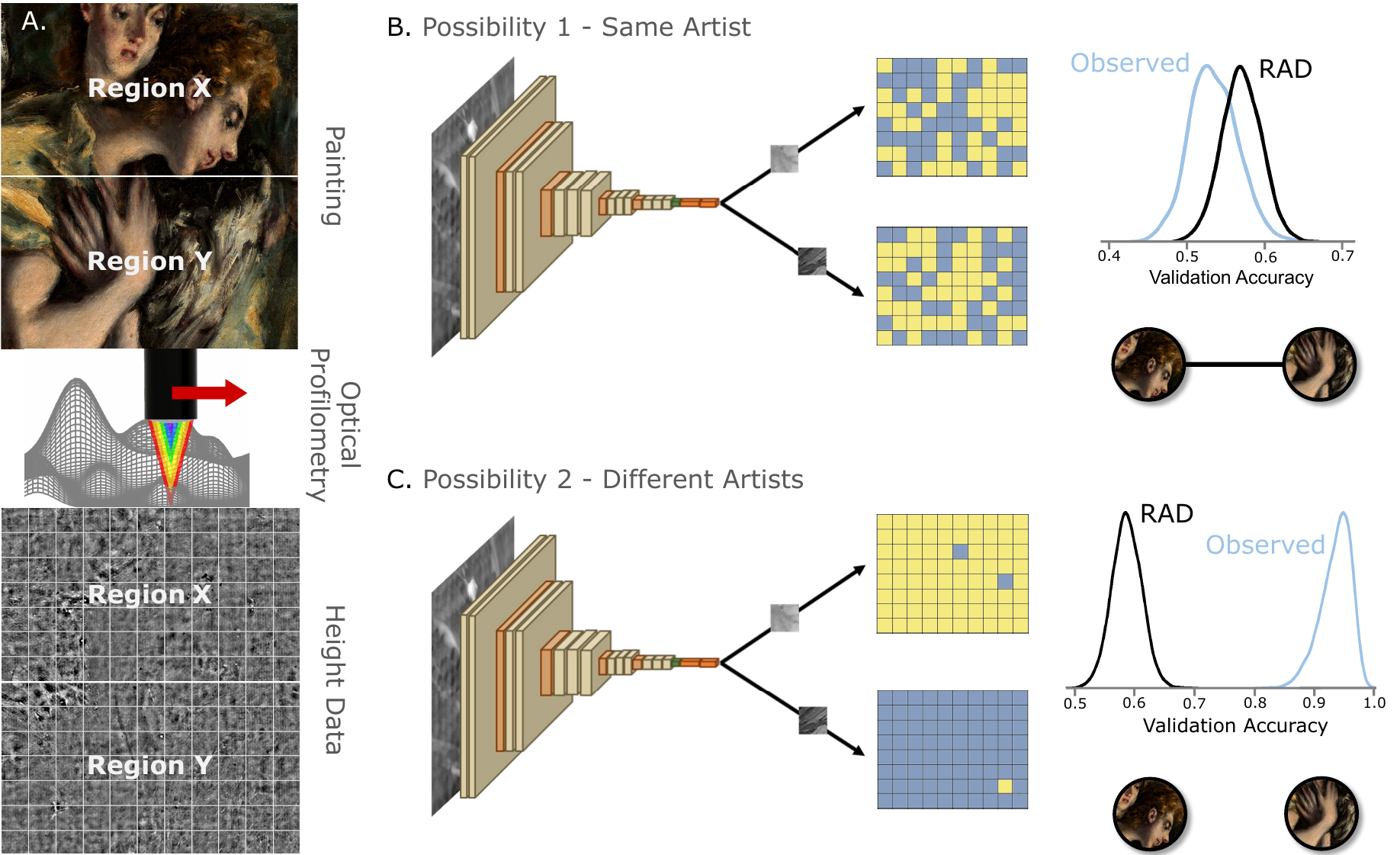}
\caption{Overview of the PATCH method: A) Paint height data is
gathered using optical profilometry, generating a 16-bit image file
encoding the height values, which is subsequently divided into
1x1 cm$^2$ patches. B, C) A CNN is repeatedly trained
to sort patches of two images, recording its accuracy as a percentage of
patches correctly sorted from the test set. This distribution of
observed accuracies is compared to the random assignment distribution
(RAD) of accuracies if the patches were sorted at random. If the
observed accuracy is no better than the RAD accuracy, then the images
were likely painted by the same artist, whereas if the observed accuracy
is better than expected by chance, the images were likely painted by
different artists.}
\end{figure}

The first phase was developed on a set of paintings by nine known student artists. Each artist individually painted three paintings, all using the same tools, materials, and subject (a photograph of lilies), as described previously\cite{ji2021discerning}. While the artists were instructed to realistically depict the subject, choices of style and technique were left up to the artists. One artist was only able to produce two paintings, and one painting was removed from the sample due to the optical profilometry data being corrupted, for a total of 25 paintings. Topographical information for each painting was recorded using high
resolution spectral confocal optical profilometry with a spatial
resolution of 50 microns, and a height repeatability of 200 nm. The resulting data were processed to remove any large-scale warping of the canvas by subtracting a mean-filtered version of the height map (with filter radius of 0.5 cm)\cite{ji2021discerning}. Because in this dataset the paintings were small (12 cm $\times$ 15 cm), each ``region'' chosen for the pairwise training consists of an entire painting. The region is then subdivided into 1 cm$^2$ square patches, which corresponds to the size where the earlier supervised machine learning analysis~\cite{ji2021discerning} was able to achieve maximum classification accuracy. This patch size is big enough to contain significant information about brushstrokes, while yielding sufficiently large datasets for training (in this case 180 patches per painting).

The network architecture chosen was VGG-16, implemented in TensorFlow
and pretrained on the ImageNet dataset. Given the computational
complexity of the training---the combinatorics of many region-to-region
comparisons, repeated over many folds, described in detail below---the
relatively light-weight but accurate VGG-16 architecture was ideal for
the PATCH approach (see the \emph{Supplementary Materials} for more information on hyperparameter settings and optimization). In order to measure the ability of the network to learn to distinguish between two paintings or regions of a painting, we ran 26
training folds. Each fold comprised the following steps: 1
cm$^2$ patches from each painting/region were randomly
selected, with replacement. Sampling with replacement creates a
bootstrapping effect that drives down the success rate of same-artist
pairs and drives up the success rate for different-artist pairs. To eliminate the influence of directional elements, the corners of each patch were trimmed to create an octagon, and the patch was randomly rotated to one of eight possible orientations. The network was then trained for 25 epochs with validation set size set
to 30\% of patches, and the maximum validation accuracy (percentage of
validation patches correctly identified) was recorded. The distribution of the 26 maximum validation accuracies was then compared to the distribution expected if the network were assigning patches at random.

If the network has failed to learn to assign objects to their correct
classes, it should perform this task no better than if it were assigning
them randomly. We are measuring the maximum validation accuracy from 25
training epochs in each of 26 folds. Thus, the distribution expected if
the network fails to learn to distinguish between painters would be the distribution \(p_{n,k}^{\max}(m)\), the probability that you will see a maximum of \emph{m} heads (i.e., correct assignments) over \emph{k} repetitions (\emph{k} epochs) of an
experiment where you flip a fair coin \emph{n} times (the number of
patches in the test set), which we call the random assignment distribution (RAD) (see the \emph{Supplementary Materials} for an analytical derivation of this distribution).

Two decision criteria were selected to determine whether the
distribution of maximum validation accuracies indicated that the network
had failed to learn to distinguish between the input paintings: 1) if
the observed mean had a z-score less than 2 relative to the RAD mean and
2) if the largest of the observed maximum validation accuracies (the
right edge of the observed distribution) among the 26 folds was less
than an empirically determined threshold based on finite sampling from
RAD, plus 10\% to account for the effects of bootstrapping (selection
with replacement, see \emph{Supplementary Information} for more
details). Using these decision criteria, the classification performance in determining
whether the artist was the same or different was exceptional. We
identified same-artist pairs (\emph{n} = 23) and different-artist pairs
(\emph{n} = 277) with F$_1$ scores of .889 and .991 respectively (see Table S2). The PATCH algorithm substantially outperformed a statistical method based on surface roughness, and beat a variety of unsupervised and supervised clustering approaches (see details in SI Appendix C) in F$_1$ for both same-artist and different-artist pairs.

While pairwise assignment training performs exceptionally well in
identifying artists with control for materials, situations where ground truth about the artistic
practices employed is not available require some means to create the \emph{post hoc} classes to which regions of a painting will be assigned. To this end, we use network analysis. We
construct a network where same-artist pairs (or, for historical
paintings, same-practice pairs) of paintings or regions are connected by
an edge (lines connecting regions in Figure 3) and different-artist/practice pairs are unconnected.

The PATCH method is concerned with ``sameness,'' meaning we need a way
to remove spurious edges from the network. To accomplish this, we employ
an edge pruning process based on uniqueness. We assume that regions with
a high degree (number of connections) have more common features and
nodes with a low degree have more unique features. Therefore, a
connection between regions with more unique features is more likely to
be correct. We operationalize this assumption by assigning each edge a
score based on the average uniqueness of its nodes, where uniqueness is
defined as the percentage of nodes a node does not connect with:
$W_{ij} = \frac{1}{2}((\text{max} - \text{degree}_{i})/\text{max} + (\text{max} -
\text{degree}_{j})/\text{max})$ where max = the maximum possible degree
(i.e., number of connections) and degree = the degree
of one of the two nodes in question. There were two erroneous
same-artist pairs in our experimental dataset (9\% of edges in the
network), and these edges had the lowest uniqueness scores. Thus, we
trim the 9\% of edges (or more if there are ties) in a given network
with the lowest uniqueness scores.

With potentially spurious edges removed and the remaining edges
weighted equally, we then employ a community finding algorithm to
identify groups of regions that are the most similar. If a set of
regions has a large number of internal links within the set and fewer
links to outside regions, it forms a ``community'' within the
network\cite{girvan2002community,newman2004finding,fortunato2016community}, suggesting that the regions in a given
community were likely painted under different circumstances (artists,
materials, etc.) than those in other communities. To establish the
degree of difference between communities, we can characterize the
community structure of the network via measures such as modularity
($Q$)---the fraction of edges in the network that are internal to
communities minus the mean fraction in a network with the same
communities but where the edges are completely
randomized\cite{newman2004finding}. We implemented the Louvain community
finding algorithm\cite{blondel2008fast} in Gephi\cite{bastian2009gephi}.
With the resolution set to 1.0, the algorithm returns the partition of
the network with the maximum modularity. The Louvain algorithm returned
a correct partition of the experimental dataset (nine disjoint communities for nine different artists) with $Q=0.875$ (see Figure S4).

\section*{El Greco, \emph{The Baptism}, and \emph{Christ on the Cross}}

El Greco, born Domenikos Theotokópoulos (1541-1614), is regarded as a
pillar of the Renaissance in Spain and as an early progenitor of
modernism\cite{hadjinicolaou2014prophet}. El Greco became a master of icon painting
in Crete before journeying through Venice and Rome, eventually settling in Toledo,
Spain\cite{marias2013elgreco,marias2014greek}. It was there that the master married
Byzantine and Venetian motifs\cite{mann2002delno} and developed the
peculiar style for which he is most widely recognized, with its
distorted human figures, ``expressive hands''\cite{sanchezcanton1964spanish}, and
"exploitation" of "pure colors to their limits"\cite{krumrine1993color}. Two
examples of that style are \emph{The Baptism of Christ} (1624, Hospital
Tavera, Toledo, Esp., henceforth, \emph{The Baptism}), shown in Figure
2A, and \emph{Christ on the Cross with Landscape} (ca. 1600-1610,
Cleveland Museum of Art, Cleveland, OH, henceforth \emph{Christ on the
Cross}), shown in Figure 2B.

These works were chosen for this analysis because of the difference in
how art historians have characterized their authorship. \emph{The
Baptism} has long been thought to feature the work of El Greco and at
least one other artist. Primary historical evidence indicates that El
Greco began the painting under a contract with the Hospital Tavera but
retained it at his death in 1614. It was delivered to the Hospital
nearly a decade later\cite{wethey1962elgreco}, during which period art
historians have proposed the painting was finished by workshop members,
particularly the master's son, Jorge Manuel\cite{marias2013elgreco}.
Previous art historical studies have attempted to attribute regions of
\emph{The Baptism} to the master and others (illustrated in Figure S3)
largely through connoisseurship---visual analysis of artists' styles and
artistic choices\cite{scallen2004rembrandt}. El Greco himself is proposed to
have painted the entire top, with the possible exception of the robe of
the angel on the right (which was not scanned for this study due to its
poor condition)\cite{wethey1962elgreco,lopera2005elgreco}, as well as the angel in green
(bottom left), with the exception of the wings. Jorge Manuel is proposed
to have painted John the Baptist (bottom right) and the adjacent
figure in red. Lopera\cite{lopera2005elgreco} and Wethey\cite{wethey1962elgreco}
ascribe the image of Christ (bottom center) to Jorge Manuel and El
Greco, respectively, suggesting that it could represent the work of both
artists. Lopera believes that a third hand may have been involved in
rendering the faces of the angels in the background on the bottom. The
landscapes at the bottom were extended and ``transformed into a river''
by an unknown artist sometime between receipt of the painting and its
installation in the epistle-side altarpiece ca. 1660, which was not an
unusual practice at the time. Restoration of the landscape on the right
at the very bottom was performed in 1936\cite{wethey1962elgreco,lopera2005elgreco,cossio1908elgreco}. In
contrast, \emph{Christ on the Cross} has been entirely attributed to El
Greco himself. It is generally accepted that El Greco created many works
in their entirety without the aid of his workshop, including
\emph{Christ on the Cross.} Francis\cite{francis1953crucifixion} states that most
of El Greco's approximately twenty renditions of the Crucifixion were
``by his own hand,'' and notes the ``lightning dexterity of brushwork''
in this particular example.

\begin{figure}[t]
\includegraphics[width=\textwidth]{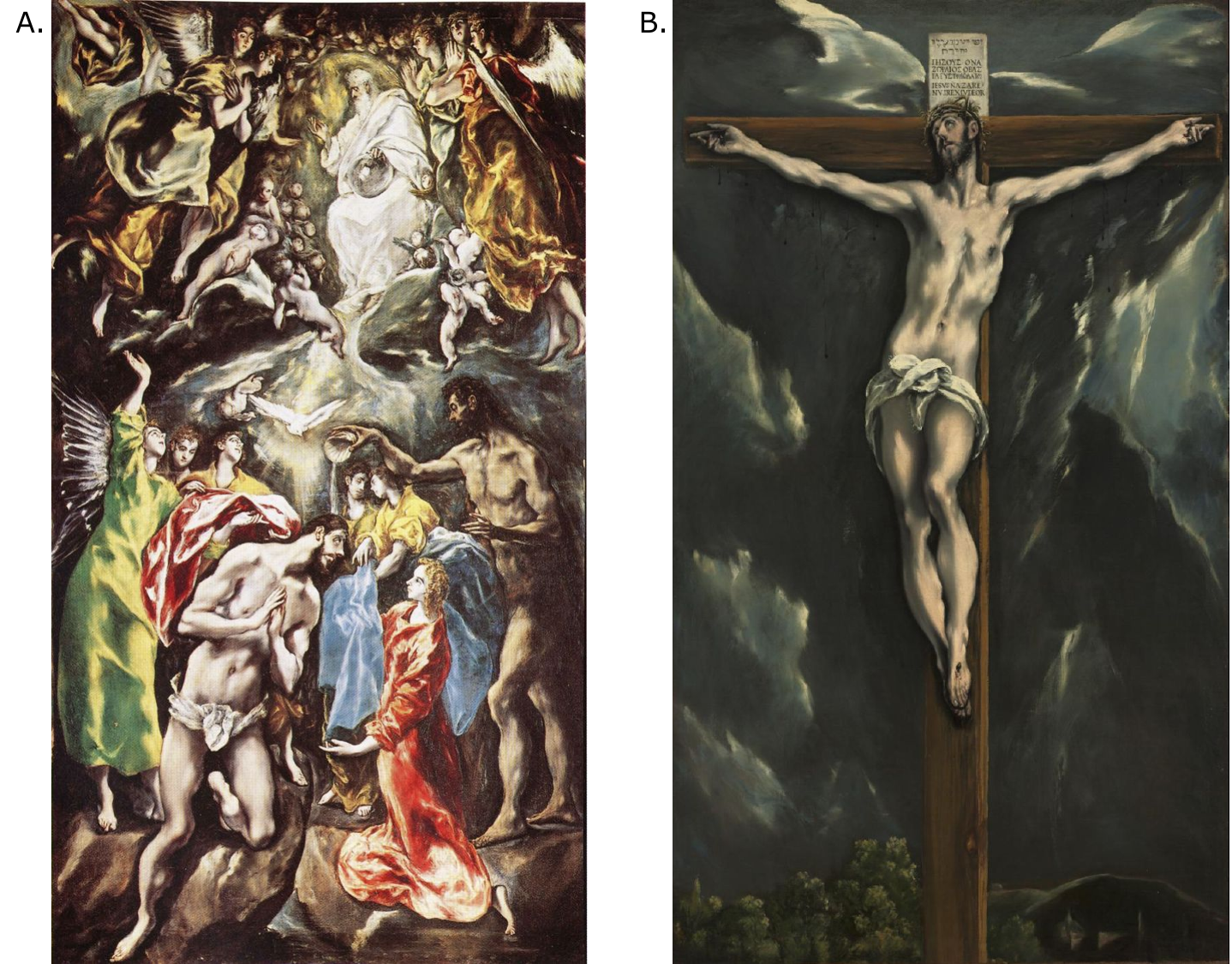}
\caption{A) El Greco's \emph{The Baptism of Christ}, begun
in approximately 1608 and suspected to have been finished by his
workshop before delivery to the Hospital Tavera in 1624. B) El Greco's
\emph{Christ on the Cross with Landscape,} ca. 1600--1610. Images are of artworks in the public domain.} 
\end{figure}

\section*{Application to El Greco's {\em The Baptism}}

Two large sections of \emph{The Baptism} were scanned \emph{in situ}
using the Factum Foundation's Lucida scanner. The Lucida scanner is a
non-contact laser triangulation scanner with 100-micron lateral
resolution and is capable of acquiring scans in situ over a large area
by stitching together 48 cm square tiles. The scanned regions used in
this study are indicated in Figure 3. The height data was processed in
the same way as for the student paintings described above, including correction for possible canvas
warping.

Regions of the topographic image of \emph{The Baptism} were selected by
hand in order to minimize variation in subject matter (e.g., faces,
robes) within each region and to constrain region size to between 180
and 540 cm$^2$. The lower bound was chosen because the
method had been developed and validated using 180-patch student paintings and the
upper bound was selected to ensure adequate representation of patches
when regions of significantly different size were compared. We avoided
selecting areas with significant cracking or damage, including the robes
of the large angels to the left and right of God, and the red swatch
held by the ``green angel'' in the bottom half of the painting. When comparing regions of
different sizes during the PATCH analysis, the sample size was set equal to the number of patches
in the smaller region.

Pairwise comparisons of all 55 regions (a total of 1485 pairwise tests)
of \emph{The Baptism} yielded a network with 356 edges, which was
reduced to 314 edges after pruning. The Louvain community-finding
algorithm\cite{blondel2008fast} with resolution set to 1.0, returned a
partition with $Q=0.341$ that featured four communities (Figure
3a). A value of $Q \ge  0.3$ is often considered evidence
of structure in real-world networks\cite{newman2004finding}.

\begin{figure}[t]
\singlespacing
\includegraphics[width=\textwidth]{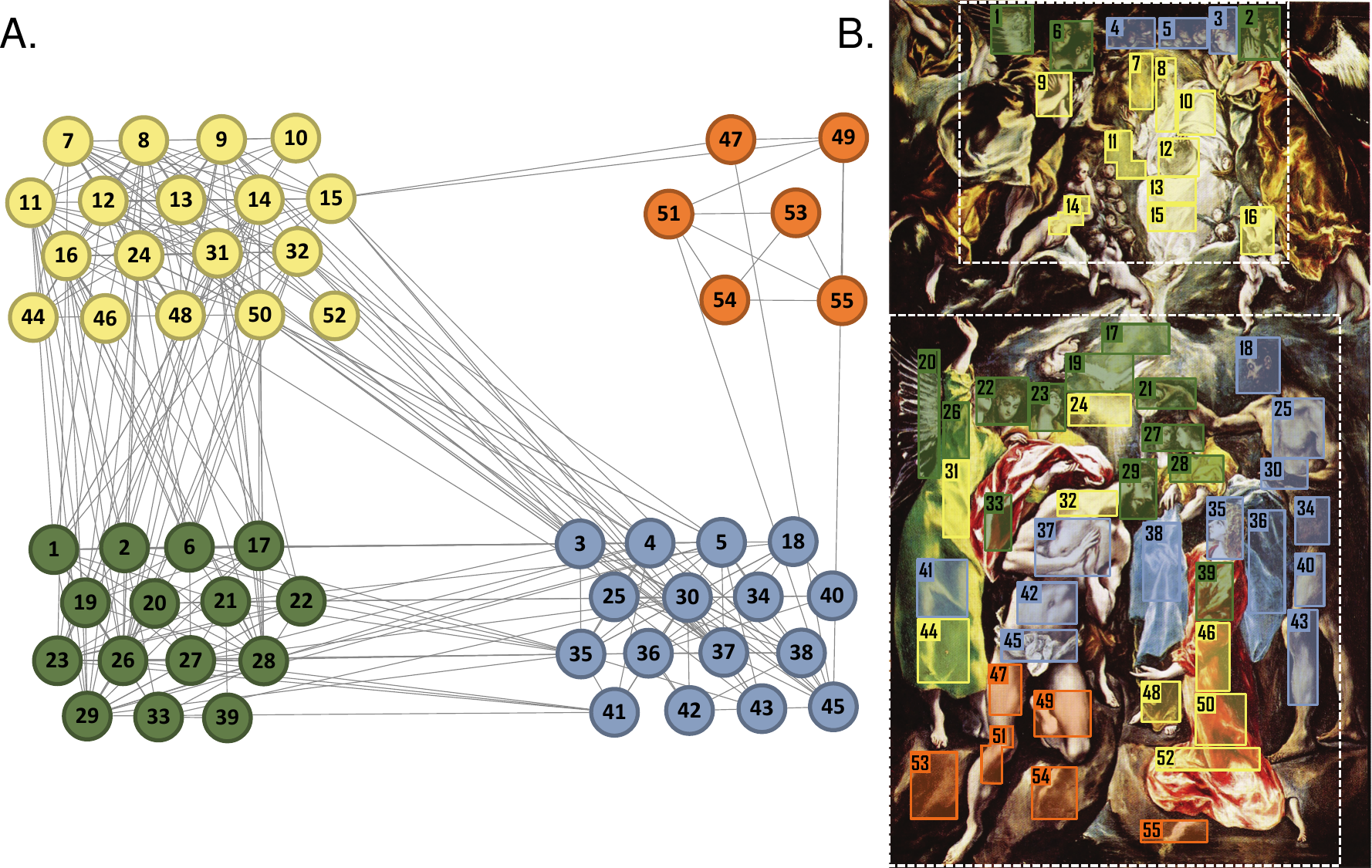}
\caption{Results of PATCH analysis on El Greco's \emph{The Baptism of Christ}
(1624, Hospital Tavera, Toledo, Esp.). Areas that were scanned are
outlined with dashed white lines. A) Network diagram showing the
maximum modularity ($Q$) partition. Four communities are present. B) the
communities mapped onto the corresponding regions of the painting that
were analyzed. Image is of artwork in the public domain.}

\end{figure}

The four communities are mapped onto the painting in Figure 3B. The
first community overlaps much of the top portion of the painting,
including the image of God the Father. The second overlaps the image of
John the Baptist and Christ's torso and hands. The third covers much of
the center of the painting, the face of Christ and the faces of many of
the background angels. The final community is confined to Jesus' legs
and the rocks at the bottom of the painting.

\section*{Application to El Greco's {\em Christ on the Cross}}

As in the case of \emph{The Baptism}, regions of the topographic image of \emph{Christ on the
Cross} were selected by hand in order to minimize variation in subject
matter and to constrain region size. Areas with apparent cracking or
damage, such as the background and trees on the lower left and the cross
beam to the immediate left of Christ's face, were not included.

Pairwise comparison of all 24 regions (276 pairwise tests) yielded a
network with 123 edges, which was reduced to 108 edges after pruning.
The Louvain community finding-algorithm\cite{blondel2008fast} at
resolution 1.0 generated a partition with $Q=0.231$ that featured
two communities. Two is the smallest possible number of communities for
a non-fully connected graph, as $Q$ compares inter- and
intra-community connections. Communities are mapped onto the painting in
Figure 4. Constituent regions of both communities are not randomly
distributed, with one community overlapping much of Christ's torso and
face, while the other community overlaps Christ's legs, the lower
portion of the cross and regions of the background.

\begin{figure}[t]
\singlespacing
\includegraphics[width=\textwidth]{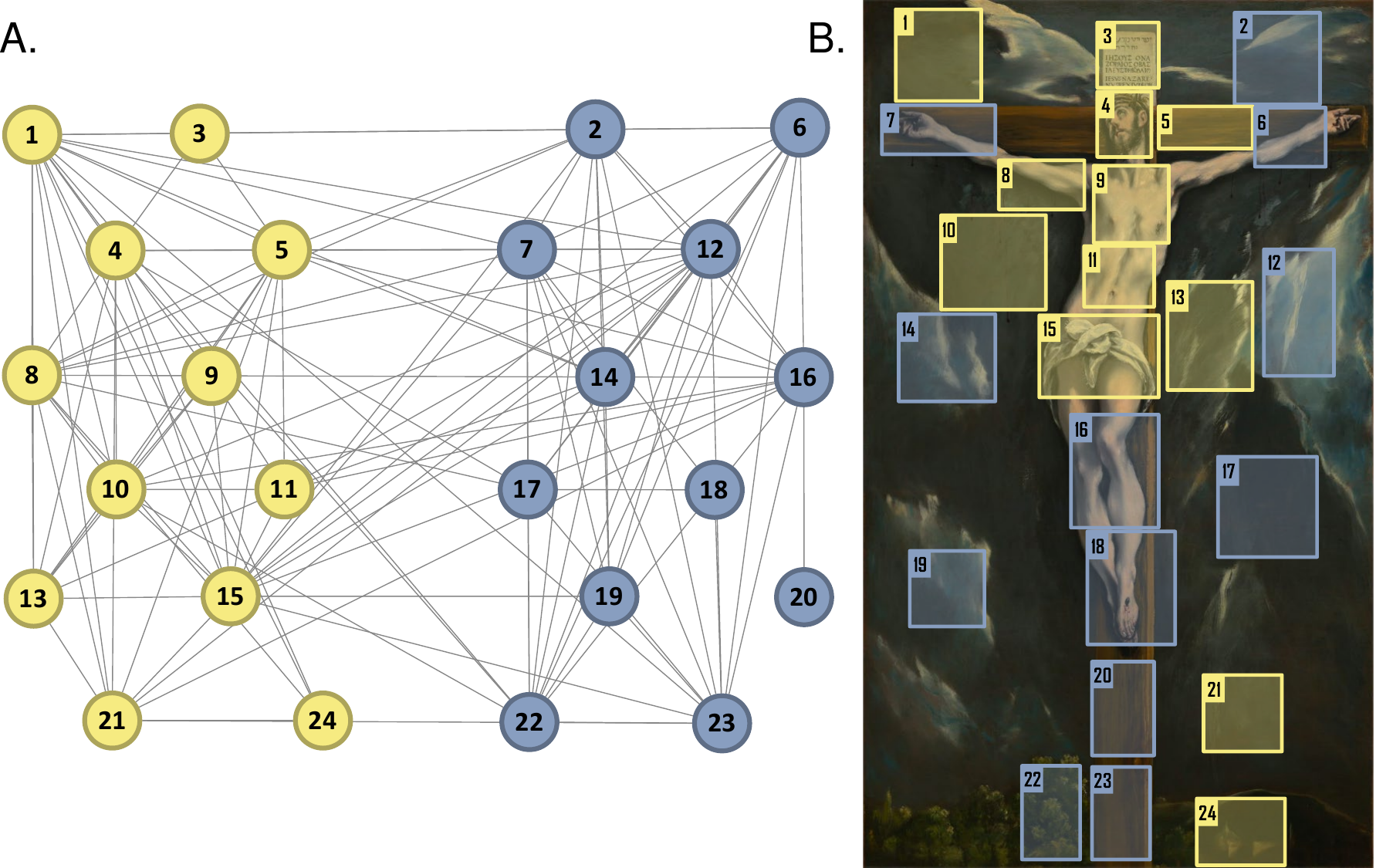}
\caption{Results of PATCH analysis on El Greco's \emph{Christ
on the Cross with Landscape.} A) Network diagram showing the maximum
modularity ($Q$) partition. Two communities are identified, though it
should be noted that this is the minimum number of communities
discoverable using a maximum-modularity partition. B) The communities
mapped onto the corresponding regions of the painting that were
analyzed. Image is of artwork in the public domain.} 
\end{figure}

\section*{Discussion}

To our knowledge, ours is the first study to demonstrate the PATCH learning
method and apply it both to paintings from a controlled experiment and
to historical paintings by a well-known artist with a workshop. Among
AI/ML-based methods designed to aid in the analysis of historical
paintings, PATCH is notable in that it requires no external, ground
truth data. Networks trained on known examples can accomplish tasks
associated with workshop practice and authorship with exceptional
accuracy. For example, Ugail and colleagues recently used transfer
learning to authenticate paintings by the Renaissance master Raphael
(Raffello Sanzio)\cite{ugail2023deep}. By combining edge detection with
typical feature extraction in a residual neural network, they were able
to recognize works by Raphael with 98\% accuracy. However, known
examples of artists' work may not be available, as exemplified by the
workshop context. Our method shows exceptional promise as a tool for ML
analysis of complex historical works where known samples of work by the
artists involved do not exist. Indeed, it may have extensive application
in other image analysis tasks where there is little to no ground truth
information available.

Further, the second phase of PATCH creates what we consider a measure of
the heterogeneity of artistic practice (HAP) within a given dataset. Communities found within a dataset represent different artists, materials, or both. The degree to which
those communities are interconnected provides information about the
possibility of shared authorship, materials, or other practices. A fully
connected network with no distinct communities has $Q = 0$. As the number
of communities increases and the interconnection between communities
decreases, modularity increases toward 1 (a completely disconnected
network has undefined modularity, but would effectively represent the
maximum heterogeneity). The networks from our three studies provide
empirical examples of lower and higher modularity networks. \emph{Christ
on the Cross} and \emph{The Baptism} have $Q$ values that cluster around
0.3, considered the threshold for evidence of structure, indicating low
heterogeneity. The network for the experimental student paintings has
much greater heterogeneity at $Q=0.875$ (see Figure S4). The
maximum modularity depends on the number of edges in the network\cite{fortunato2007resolution}, but also on the number of communities in the
network. A network with two fully connected and completely disjoint
communities has $Q = 0.5$, for example. As such, network modularity can
function as a measure of absolute observed heterogeneity (Figure 5).

\begin{figure}[t]
\centering
\includegraphics[width=0.95\textwidth]{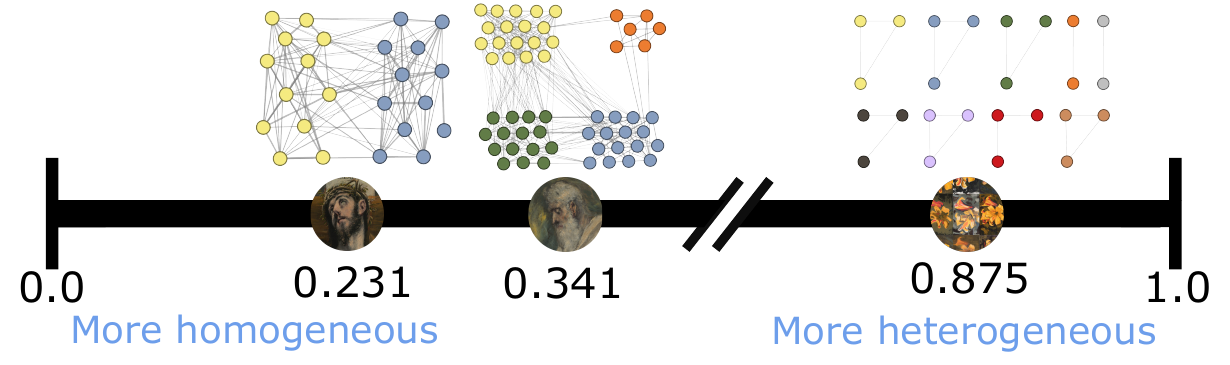}
\caption{Networks and modularity ($Q$) values for {\em Christ on the Cross}, {\em The Baptism}, and the experimental student paintings along the axis of $Q$ from
least to most heterogeneous.} 
\end{figure}

HAP informs interpretation of our analyses of \emph{Christ on the Cross}
and \emph{The Baptism}. For the former, our analysis identifies two
communities of regions, but with $Q$ below the threshold generally
accepted to indicate structure (0.231 vs 0.3). The regions in each
community do cluster in space, however, which is worth considering. One
community overlaps Christ's head and torso and surrounding areas.
Conservation files generously provided by the Cleveland Museum of Art
indicate that the figure of Christ (areas overlapped by the community in
yellow in Figure 4) contains inpainting from previous conservation,
which could create a distinguishable signal, though damage to the canvas
is another, more likely culprit. Further mapping of damage and past
conservation efforts will improve our understanding of this work.
However, the results comport with the notion of \emph{Christ on the
Cross} as the work of a single artist with some variation introduced by
artistic practice or possibly early conservation practice.

At first glance, the communities identified in our analysis of \emph{The Baptism} appear to roughly correspond to the proposed attributions by previous authors discussed above. The community in yellow in Figure 3 overlaps much of the top of the painting, previously
attributed to El Greco, while the community in blue overlaps John the
Baptist and the figure in red, assigned to Jorge Manuel. We also find a
combination of communities in the face and torso of Christ, which was
attributed to both artists. A single, small and particularly unique
community (in orange in Figure 3) overlaps the landscape at the bottom
of the painting and Christ's legs.

Overall, however, the PATCH findings suggest a different story
underlying the existing art historical interpretation. The partition of
\emph{The Baptism} has $Q = 0.341$, which is only slightly above the
threshold for evidence of structure. Nearly a third of all edges in the
network are interconnections between three of the four communities we
have identified. The yellow and green groups have 49 intercommunity
edges, blue and green 30, and blue and yellow 22. If the blue, green, and
yellow communities indeed represented three distinct artists working
with the same or similar materials, we would expect the communities to
be more insular and the modularity of the network to be substantially
higher, similar to what we observed in our controlled student painting dataset. Yet the
modularity of \emph{The Baptism} is much closer to that of \emph{Christ
on the Cross} ($Q = 0.231$), which appears to feature one artist\emph{.}
It is worth noting that, as shown in the analysis of student paintings,
the CNN is particularly proficient at recognizing different-artist pairs
and is more likely to incorrectly label a same-artist pair as different
than the converse. Thus, the intercommunity edges in \emph{The Baptism}
network that survived our initial pruning are not liable to be spurious.

This suggests that some unifying factor connects these three
communities. One possibility is a single artist working with different brushes. Renaissance artists used brushes made from the bristles or hair of several different mammal species, variation in the characteristics (e.g., coarseness) of which could affect the deposition of paint \cite{gettens2012painting,palomino1986pictorial,cennini2018book} (see \emph{Supplementary Materials} for more information). The evidence could also indicate the work of one artist with changing style or technique over time. Interestingly, Biannuci and
colleagues\cite{bianucci2017historical} present evidence that El Greco suffered a
series of ischemic events, one in the 1590s and a second in 1608 (the
year the contract for the altarpieces for the Hospital Tavera was
initiated), ``resulting in progressive disabilities with fluctuating
course characterized by temporary improvements and worsening before his
death.'' The regions that were previously identified as the work of
different individual painters could conceivably represent variation in
the master's individual style (however this may have arisen) over the
course of his final years. Investigation of changes in style and
technique and their effect on PATCH analysis will be an important
direction for future research.

PATCH has the capability to make a substantial contribution to research
as a complement to existing art historical methods. The accuracy of the
method has been demonstrated in the validation on student paintings. Our analyses of \emph{Christ on the Cross} and \emph{The Baptism} show the true
potential of PATCH by contributing important new information to art
historical scholarship on the creation of objects in El Greco's workshop
in Toledo. In the analysis of \emph{The Baptism}, the fact that the orange group,
which overlaps a region known to have been altered after the delivery of
the painting, is so disconnected from the remainder of the communities (see \emph{Supplementary Materials} for further analyses) speaks to the method's capability of recognizing a unique contribution or substantial departure from the technique used in the rest of the
painting. That the other three communities roughly correspond to art
historical attributions suggests PATCH possesses a kind of semantic
fidelity\cite{takagi2023high}---our method recognizes areas that a trained art historian would classify as different. However, the
connections between those areas revealed by PATCH also call those attributions
into question. Computer vision allows us to view the surface of a painting at a
different scale. Previous analyses indicate that features as small as
the diameter of a single paint brush bristle (length scales of 0.2--0.4
mm) may be integral to the network's identification of an
artist\cite{ji2021discerning}. By analyzing spatial correlations at this
microscopic scale, the network may be revealing aspects of the physics
of paint application and of the physiology of hand movements. PATCH adds
micro-scale features to the macro (brushstrokes) and meta-scale
(historical and material) data, allowing for a full-scale analysis and
the discovery of heretofore unseen evidence of the processes by which
paintings were created.

Future research with PATCH will focus on teasing apart the influences of materials and artists on the heterogeneity of artistic practice. Expansion of PATCH applications to include \emph{intra-}painting comparisons and other art historical applications will be important as well. However, uses of PATCH are not limited to those contexts: it provides a general-purpose approach to assessing regional similarity in images or data that can be encoded in image-like arrays.
As such, PATCH is a promising method for applications where supervised learning is impossible (because the ''ground truth'' is unknown) and alternative unsupervised approaches are ineffective, especially for \textit{a posteriori} cluster number. Potential applications span the physical and social sciences, and include image segmentation tasks in applications such as medical imaging\cite{raza2021tour}, agricultural remote sensing\cite{khanal2020remote}, urban development and design\cite{francis2023unsupervised}, and microstructural analysis, as well as anomaly detection in a range of manufacturing contexts\cite{kharitonov2022comparative,liu2024deep}, among others.

\section{Data availability}
The dataset used to develop and validate the PATCH method is available on OSF at \url{osf.io/a5xnh/}. Height data from \emph{The Baptism} and \emph{Christ on the Cross} are the property of the institutions that own those paintings: the Fundación Casa Ducal de Medinaceli and the Cleveland Museum of Art, respectively.

\section{Code availability}
The code used to run the PATCH algorithm is available on GitHub: \url{https://github.com/hincz-lab/PATCH}.

\singlespacing

\bibliographystyle{unsrt}
\bibliography{citations}

\section*{Acknowledgments}

The authors would like to thank Fundación Casa Ducal de Medinaceli
for providing data for \emph{The Baptism} and Sr. Juan Manuel
Albendea Solis for his feedback on our initial draft, which helped us reevaluate
our conclusions and improve the manuscript. We would also like to thank
the Cleveland Museum of Art for use of the data from \emph{Christ on the
Cross}, and the Lapis Senior Conservator of Paintings and Head of
Paintings Conservation, Dean Yoder, for his help and feedback. The
authors acknowledge support from the Expanding Horizons
Initiative of Case Western Reserve University (CWRU) College of Arts
and Sciences and the Jack, Joseph and Morton Mandel Foundation.
Computational support was provided by the CWRU High Performance
Computing cluster and a workstation donated through the Dell Seed
Program.

\section*{Author contributions}
AV conceived the PATCH method; LNS and MSM conceived the initial project; AV, MH, KS, MSM developed the method; AI oversaw creation of paintings; ALo, CB, and ITM performed scans; AV and MM wrote PATCH code; AV, ALi, MM and MH performed experiments and analyses; AV, MH, KS, ESB, CP, ALi, and LNS wrote the manuscript.  

\section*{Ethics declarations}
Competing interests: The authors declare no competing interests.
\section*{Supplementary information}
Supplementary Information is available for this paper.
\section*{Correspondence}
Correspondence and requests for materials should be addressed to AV.

\newpage
\setcounter{page}{1}
\setcounter{table}{0}
\setcounter{figure}{0}

\renewcommand{\figurename}{\fontfamily{phv}\fontsize{9}{12}\selectfont Figure}
\renewcommand{\thefigure}{S\arabic{figure}}
\renewcommand{\tablename}{\fontfamily{phv}\fontsize{9}{12}\selectfont Table}
\renewcommand{\thetable}{S\arabic{table}}
\renewcommand{\thepage}{S\arabic{page}}

\renewcommand{\Authfont}{\fontfamily{phv}\fontsize{11}{13}\fontfamily{phv}\selectfont}
 \renewcommand{\Affilfont}{\fontfamily{phv}\fontsize{10}{12}\selectfont}

\fontfamily{phv}\selectfont
    
\begin{center}
\LARGE Supplemental Materials:\\
``PATCH: a deep learning method to assess heterogeneity of artistic practice in historical paintings''
\end{center}

\section*{\fontfamily{phv}\selectfont Appendix A: Analytical derivation of the random assignment distribution}

In order to determine whether the network has learned to correctly sort patches of both images, we generated the distribution of correct assignments expected if the network were assigning each patch to a class at random (as though it were simply flipping a coin). The network is trained for 26 folds of 25 epochs each. In each epoch, the network is trained on 70\% of the patches and tested on the remaining 30\% (there are equal numbers of patches from each image). The percentage of this test set that is assigned correctly is called the “validation accuracy.” At the end of each fold, the maximum validation accuracy is recorded, the dataset is resampled, and a new fold begins. After all folds are completed, we have a distribution of 26 maximum validation accuracies that represent how well the network has learned to distinguish between the two images in each of its attempts. The derivation of this expected distribution is as follows:

The goal is to find the distribution $p^{\text{max}}_{n,k}(m)$, the probability that you will see a maximum of $m$ heads over $k$ repetitions (``$k$ epochs'') of an experiment where you flip a fair coin $n$ times. We start with the distribution of finding $m$ heads in a single experiment, which is the binomial distribution for a fair coin,
\begin{equation}
    p_n(m) = 2^{-n} \binom{n}{m},
\end{equation}
where $\binom{n}{m} \equiv \frac{n!}{m!(n-m)!}$ is the binomial coefficient. We also define a tail distribution, which is the cumulative probability that you would find $m$ or more coin flips in a single experiment:
\begin{equation}
    p^{\text{tail}}_n(m) = \sum_{j=m}^{n} p_n(j) = \binom{n}{m} {}_{2}F_1(m,n+1,m+1; -1).
\end{equation}
Here ${}_2 F_1(a,b,c;z)$ is a hypergeometric function, defined by the series:
\begin{equation}
{}_2 F_1(a,b,c;z) = \sum_{j=0}^\infty\frac{ (a)_j (b)_j}{ (c)_j}\frac{z^j}{j!},
\end{equation}
where $(x)_j \equiv x (x+1) \cdots (x+j-1)$ for $j>0$ and $(x)_0 \equiv 1$.  There are other equivalent ways of expressing the tail probability of a binomial distribution (using for example incomplete beta functions), but all of them are similarly complicated.

If we repeat the experiment $k$ times, the probability that every single experiment had fewer than $m$ heads is:
\begin{equation}
    U_{n,k}(m) = (1- p^{\text{tail}}_n(m))^k.
\end{equation}
The difference between consecutive values of this probability, $U_{n,k}(m+1) - U_{n,k}(m)$, is the probability that we would find a maximum exactly equal to $m$ over $k$ repetitions.  Hence we get our final answer:
\begin{equation}
    p^{\text{max}}_{n,k}(m) = U_{n,k}(m+1) - U_{n,k}(m).
\end{equation}
It is easy to check that this probability is properly normalized:
\begin{equation}
    \sum_{m=0}^n p^{\text{max}}_{n,k}(m) = U_{n,k}(n+1) - U_{n,k}(0) = 1.
\end{equation}

In practice, we compared the mean and maximum of the observed distribution to the random assignment distribution derived above. However, the theoretical maximum of the distribution is \textit{m} = \textit{n}. Given enough tries (infinite sampling), a random assignment engine would eventually achieve perfect accuracy. Thus, we developed an empirical “maximum” value based on finite sampling using the image size of our student paintings (180 patches, 360 total, \textit{n} = 108 patches in the test set). We created a simple coin-flipping script that gathered the maximum “validation accuracies” (i.e., percentage of flips that land heads) for 25x108 flips and ran 250,000 iterations of this process. This generated a maximum value of \textit{m} = 80 “correct assignments,” or 74.074\%. The probability of 80 correct assignments is ~0.000004676581 (truncated for clarity). As the distribution is a probability mass function based on discrete events, for tests with larger numbers of patches, we used the number of correct assignments with a probability closest to this probability, even if it was less probable. 

\section*{\fontfamily{phv}\selectfont Appendix B: Hyperparameter Settings and Optimization}
The first phase of PATCH analysis, pairwise assignment training, was performed using VGG16 architecture pretrained on the Imagenet database. Output from VGG-16 was pooled and flattened and
passed through two dense layers with dropout rates of 0.25. The learning rate was set to 0.0001 and the batch size was set to 32 patches. 

The model hyperparameters for the neural network artistic practice discriminator model are optimized with a commercially available Bayesian hyperparameter optimization package built around the tree parzen optimization algorithm~\cite{bergstra2015hyperopt}. 
The optimization loss is based on the model performance after initial training on a restricted subset of the training data, because using the full dataset for this initial optimization step is computationally infeasible. In particular each model was trained for: 6 folds, 10 epochs, and patch size 200 on a reduced student painting dataset consisting of 5 paintings with 4 same artist combinations and 6 different artist combinations. Training loss was calculated as the RMSE distance between the predicted artist similarity and the ground truth, averaged over all painting combinations. The restricted training scheme was employed to reduce the total optimization time on finite computational resources. The considered hyperparameters and ranges are shown in Table S1. The model structure consists of a variable number of dense, fully-connected layers following the VGG16 model, where the number of nodes per layer decreases linearly from a variable maximum at the first dense layer to 2 at the output layer. Maximum posterior estimation was performed on the generated set of all trained models (130 instances) to predict the optimal hyperparameter set. Final optimization of the model shown in the manuscript utilized a full training scheme on the full dataset, as described in the manuscript, and is performed by hand. This is necessary since the full training scheme leads to unreasonably long convergence time in the optimization algorithm. The final hyperparameters align closely with the maximum posterior estimation result, as shown in Figure S1. Note that the largest deviation between the optimized and maximum posterior hyperparameters occurs for the learning rate, which is expected since the longer number of epochs in the full training process allows the model to converge with a smaller learning rate.

\begin{table}
\fontfamily{phv}\selectfont
\renewcommand{\arraystretch}{1.05}
\centering
\begin{tabular}{|l|c|c|c|c|}
\hline
\textbf{Hyperparameter}&\textbf{Min search val.}&\textbf{Max search val.}&\textbf{Max posterior}&\textbf{Final optimization}\\
\hline
Learning rate&0.0001&0.02&0.0025&0.0001\\
Dense layers&1&4&1&1\\
Max nodes&25&200&97&64\\
Dropout rate&0.0&0.25&0.20&0.25\\
L2 reg. rate&0.0001&0.13&0.0007&0.001\\ 
\hline
\end{tabular}
\caption{\fontfamily{phv}\selectfont Optimization ranges for the searched hyperparameters.}
\label{table:S1}

\end{table}

\begin{figure}[] \centering \includegraphics[width=\textwidth]{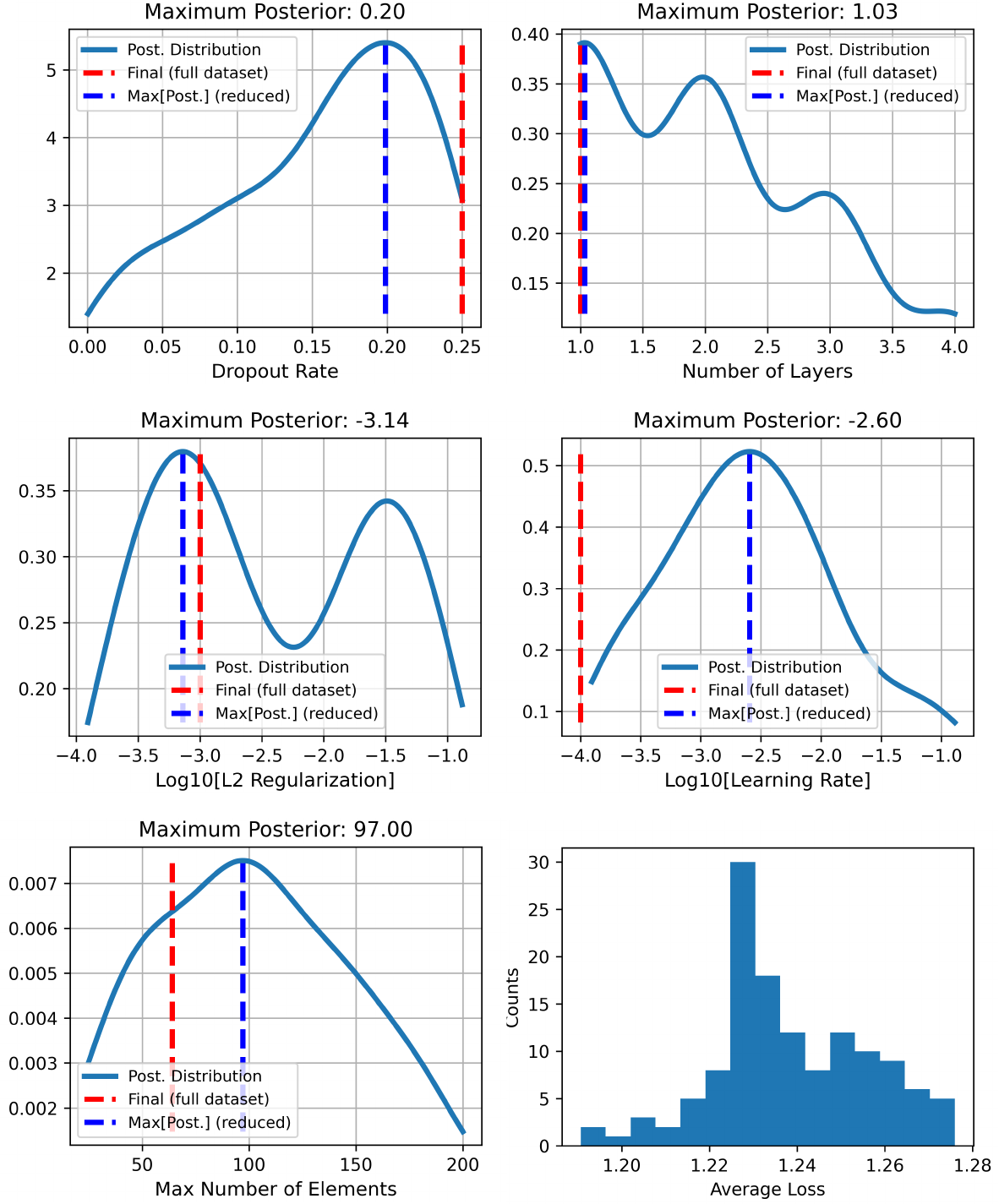}
\caption{Results of the hyperparameter optimization on a reduced dataset compared to the final model trained on the full dataset, shown in the manuscript. The posterior distribution over all optimization training is shown for each hyperparameter (blue curve), with the maximum posterior from the optimization process (blue dashed line) and the final model hyperparameters (red dashed line) indicated for comparison. The final model hyperparameters align closely with the optimization maximum posterior in most cases. The distribution of the loss for all trained model instances in the optimization is shown in the bottom right panel.}
\end{figure}

\newpage

\section*{\fontfamily{phv}\selectfont Appendix C: PATCH in Comparison to Alternative Methods}

The PATCH algorithm outperformed alternative methods (described below) for the student painting dataset, identifying same-artist pairs (\emph{n} = 23) and different-artist pairs (\emph{n} = 277) with excellent classification performance, summarized in Table S2. In the table, precision and recall are calculated by treating each category (same-artist pairs or different-artist pairs) as ``positives'' in turn. For example, for the same-artist pairs, the true positives TP $=$ number of same-artist pairs classified as same, false positives FP $=$ number of different-artist pairs classified as same, and false negatives FN $=$ number of same-artist pairs classified as different. The precision = $\text{TP}/(\text{TP}+\text{FP})$, the recall = $\text{TP}/(\text{TP}+\text{FN})$, and the F$_1$ score is the harmonic mean of precision and recall. The definitions for the different-artist pair category are as above, but with ``same'' and ``different'' interchanged. The last three columns of the table show averages of the same-artist and different-artist columns for precision, recall, and F$_1$. The PATCH method was superior to the tested alternatives in the majority of metrics, notably achieving the highest F$_1$ scores for same-artist pairs and different-artist pairs. The closest competitor, supervised UMAP with a Fourier feature extractor, achieved only marginally higher precision for same-artist pairs and recall for different-artist pairs, beating PATCH by 2.9\% and 0.3\% respectively. However it performed substantially worse on same-artist recall (0.652 versus 0.87), thus scoring lower on F$_1$. That method's marginally increased precision is therefore outweighed by its inability to identify same-artist pairs (15/23, \~65\%). In the following we describe the alternative methods we tested.
\vspace{1em}

\begin{table}
\centering
{\footnotesize \centering
\renewcommand{\arraystretch}{1.1}
\fontfamily{phv}\selectfont
\begin{tabular}{|l|ccc|ccc|ccc|} 
\hline
  & \multicolumn{3}{c|}{Same-artist pairs ($n=23$)} & \multicolumn{3}{c|}{Different-artist pairs ($n=277$)} & \multicolumn{3}{c|}{Average}\\
Method & Precision & Recall & F$_1$ & Precision & Recall & F$_1$ & Precision & Recall & F$_1$\\
\hline
\textbf{PATCH} & 0.909 & \textbf{0.87} & \textbf{0.889} & \textbf{0.989} & 0.993 & \textbf{0.991} & 0.949 & \textbf{0.931} & \textbf{0.940}\\ 
Surface roughness & 0.250&0.826&0.384 & 0.982&0.794&0.878 & 0.616 & 0.810 & 0.631\\
Fourier / UMAP $d=4$ &0.257 & 0.783 & 0.387 & 0.978 & 0.812 & 0.888 & 0.618 & 0.797 & 0.637 \\
Fourier / sUMAP $d=4$ & 0.8 & 0.522 & 0.632 & 0.961 & 0.989 & 0.975 & 0.881 & 0.755 & 0.803 \\
Fourier / UMAP $d=8$ & 0.254 & 0.652 & 0.366 & 0.967 & 0.841 & 0.9 & 0.611 & 0.747 & 0.633 \\
Fourier / sUMAP $d=8$ & \textbf{0.938} & 0.652 & 0.769 & 0.972 & \textbf{0.996} & 0.984 & \textbf{0.955} & 0.824 & 0.877 \\
Fourier / UMAP $d=12$ & 0.274 & 0.739 & 0.4 & 0.975 & 0.838 & 0.901 & 0.624 & 0.788 & 0.65 \\
Fourier / sUMAP $d=12$ & 0.688 & 0.478 & 0.564 & 0.958 & 0.982 & 0.97 & 0.823 & 0.73 & 0.767\\
CNN / UMAP $d=4$ & 0.197 & 0.565 & 0.292 & 0.957 & 0.809 & 0.877 & 0.577 & 0.687 & 0.584 \\
CNN / sUMAP $d=4$ & 0.37 & 0.739 & 0.493 & 0.976 & 0.895 & 0.934 & 0.673 & 0.817 & 0.713 \\
CNN / UMAP $d=8$ & 0.23 & 0.609 & 0.333 & 0.962 & 0.83 & 0.891 & 0.596 & 0.72 & 0.612 \\
CNN / sUMAP $d=8$ & 0.34 & 0.739 & 0.466 & 0.976 & 0.881 & 0.926 & 0.658 & 0.81 & 0.696 \\
CNN / UMAP $d=12$ & 0.239 & 0.696 & 0.356 & 0.97 & 0.816 & 0.886 & 0.604 & 0.756 & 0.621 \\
CNN / sUMAP $d=12$ & 0.348 & 0.696 & 0.464 & 0.972 & 0.892 & 0.93 & 0.66 & 0.794 & 0.697\\
\hline
\end{tabular}}
\caption{Classification metrics on the student painting dataset for the PATCH algorithm versus alternative methods described in Appendix C. The definitions of precision, recall and F$_1$ are given in the appendix text for the same-artist pairs and different-artist pairs categories. The last three columns are averages of the precision, recall, and F$_1$ in the same-artist and different-artist columns. The second row, ``Surface roughness'' refers to a statistical method that sorts patches by their standard deviation of height values. The remaining rows correspond to results from unsupervised (UMAP) or supervised (sUMAP) classification based on the UMAP dimensional reduction algorithm. In the first term of the method label ``Fourier'' or ``CNN'' refers to the type of feature extractor, and in the second term $d$ is the dimension of the vector output by UMAP.}
\end{table}

\noindent\textit{Surface roughness method}\\

Pairwise comparison of large numbers of regions via PATCH is computationally expensive, so a simpler method would be preferable if it were equally accurate. One simple way of distinguishing between regions of a painting using topography would be to compare the “roughness” of the surfaces. We applied a roughness-based method to the student paintings to determine its accuracy relative to the PATCH method. We used the standard deviation of height values as a measure of roughness, and calculated the standard deviation of height values for each 1x1 cm$^2$ patch of each student painting. We then compared paintings pairwise using the pairwise Wilcoxon rank sum test implemented in R using the stats package. A significant result on the pairwise Wilcoxon test means the distributions (specifically the medians) are significantly different, which in this context would indicate a different-artist pair. Thus, same-artist pairs would be tests where the null hypothesis could not be rejected. Using the Bonferroni correction for multiple tests with $\alpha$=0.01, any pairwise comparison with a \textit{p}-value greater than 0.00003 was considered a same-artist pair. This method resulted in F$_1$ scores of 0.384 for same-artist pairs and 0.878 for different-artist pairs, substantially worse than PATCH. This also indicates that PATCH is not simply sorting patches of images based on their roughness, but using more complex characteristics of the height data for classification.

\vspace{1em}

\noindent\textit{Unsupervised UMAP methods}\\

As described in the main text, PATCH achieves unsupervised ends by supervised means. How would PATCH fair against fully unsupervised learning on the surface topography, using clustering of dimension-reduced feature vectors for each patch? To facilitate this comparison, we developed an unsupervised pairwise region analysis method, in the spirit of the PATCH approach. We used the student paintings for validation, dividing them up into 1x1 cm$^2$ (200 x 200 px) patches, giving 180 patches per painting. We then took every possible pair of paintings, to determine whether each pair could be distinguished (marked as “different artist”) or not (marked as “same artist”). For a given pair of paintings (360 patches total) we ran each patch through the following unsupervised workflow:

\begin{itemize}

\item {\em Feature extraction:} each patch was converted to a feature vector using one of two feature extractors: i) a 2D Fourier transform, which created a Fourier domain 200x200 patch with the origin in the center. We took the absolute values of the upper half of the domain (the bottom was identical by symmetry), leaving out the 13 central columns. These columns correspond to the largest wavelengths (comparable to the patch size), and deemphasizing them puts the focus on brushstroke details at smaller wavelengths. We flattened the absolute values into a feature vector of size 17578. Other parameter choices gave comparable or worse final performance; ii) a feature extractor based on the VGG16 convolutional neural network (CNN) that had been trained to classify height data in our earlier student painting controlled experiment~\cite{ji2021discerning}. 
The bottom layers of the network were cut out, and the output was pooled and flattened to yield a feature vector of size 512. These two types of feature extractors are labeled as ``Fourier'' or ``CNN'' respectively in Table S2.

\item {\em Dimension reduction:} to allow eventual clustering of the feature vectors, we used the UMAP dimension reduction algorithm~\cite{mcinnes2018umap}
 with default parameters, to map the full feature vectors onto $d$-dimensional vectors. Three different choices of $d$ were tested: $d=4$, $8$, and $12$.

\item {\em Clustering:} Finally, we used the flat clustering module of HDBSCAN~\cite{mcinnes2017hdbscan} 
(minimum cluster size = 25, cluster selection method = ‘leaf’) to force the algorithm to group the $d$-dimensional vectors into two clusters. 

\end{itemize}

If there was a statistically significant difference in the numbers of patches assigned to each cluster among the two paintings (chi-squared test with \textit{p} $<$ 0.05), then the pair of paintings were marked as different artist. For example, if the labeled patches in one painting were mostly assigned to cluster 1, and the labeled patches in the other painting to cluster 2, this would be taken as evidence that different artists were involved. Otherwise (i.e. if cluster assignments were fairly evenly distributed among the two paintings) the pair was marked as same artist. The algorithm occasionally failed to find two clusters at all, which we also treated as evidence that the pair was made by the same artist. The resulting classification metrics, regardless of feature extractor or dimension $d$, were quite similar to those of the surface roughness approach described above. Changing HDBSCAN parameters like minimum cluster size and the cluster selection method did not substantially alter the performance, and in all cases the unsupervised approach fared worse than the PATCH method.\\

\noindent\textit{Supervised UMAP (sUMAP) methods}\\

The final set of methods we tested were supervised variants of the UMAP procedure above. These take advantage of the fact that UMAP can be nudged to provide more distinct separation of clusters by giving it class labels. The approach is the same as earlier, with three differences: 1) 70\% of the patches in each of the two paintings are randomly selected to be training data. These patches are given distinct labels by painting (i.e. painting 1 versus painting 2), which are input to the UMAP algorithm along with the patches. 2) The remaining 30\% of the patches are used to test whether the approach can successfully distinguish paintings by putting each painting in a distinct cluster. As before, if a statistically significant difference exists in the number of test patches assigned to each cluster among the two paintings (chi-squared test with $p < 0.05$) then the pair is marked as different-artist. 3) Because of the stochastic nature of training / test split, the analysis was repeated three independent times, and where there was disagreement we used the majority outcome of the three runs to decide on same-artist or different-artist for each pair.

Both Fourier and CNN feature extractors were tested with this approach, as well as different output dimensions $d$. The supervised component did improve the results, and as seen in Table S2 same-artist precision and different-artist recall were slightly better than PATCH (by 2.9\% and 0.3\% respectively). However the same-artist recall was relatively poor (0.652 for this method versus 0.87 for PATCH). Hence overall PATCH was still superior in classification, as evidenced by its higher F$_1$ scores for both same-artist and different-artist pairs. Additionally, the computational complexity of this approach (unlike the simpler surface roughness and unsupervised UMAP methods) was comparable to implementing PATCH, so that there were no substantial performance benefits.

\section*{\fontfamily{phv}\selectfont Appendix D: Analysis of PATCH-generated Communities}

Further analysis of the community structure in the PATCH generated networks can provide additional insight into the community cohesiveness and distinction beyond the modularity index. These properties are quantified by the average community degree, broadly defined as the weighted and normalized sum of all edges between nodes within two community distinctions. This analysis can provide additional insights into the cohesiveness of artistic practice within a single community, and the relative similarity of practice between multiple communities.

The internal degree ($k_{int}$) is defined as the weighted sum of all edges ($w_{ij}$) between nodes in the same community ($C$), normalized by the theoretical maximum number of edges for $n$ available nodes:
\begin{equation}
    k_{int} = \frac{2}{n(n-1)} \sum_{\substack{i,j \in C}} w_{ij}.
\end{equation}
The internal degree can be taken as a quantified metric of the intra-community cohesiveness, related to the variability of textures within an identified community.
The external degree ($k_{ext(C,C')}$) between two communities is similarly defined as the weighted sum of all edges $w_{ij}$ between nodes in distinct communities ($C$, $C'$), normalized by the theoretical maximum number of edges for $n$ and $n'$ available nodes in the respective communities:
\begin{equation}
    k_{ext(C,C')} = \frac{1}{n~n'} \sum_{\substack{i\in C,~ j\in C'}} w_{ij}.
\end{equation}
The external degree can be taken as a quantified metric for the similarity between two separate communities, related to the distinctiveness of textures between them. Note that due to the prior identification of distinct communities, the external degree is expected to be much lower than the internal.  
Finally, the total external degree for a community ($C$) is defined as the external degree between all nodes internal to ($n$) and external to ($n'$) the community, again normalized by the theoretical maximum number of edges:
\begin{equation}
    k_{ext} = \frac{1}{n~n'} \sum_{\substack{i\in C,~ j\notin C}} w_{ij}.
\end{equation}
The total external degree is a quantified metric describing the relative isolation of a single community from the remainder of the network, related to the overall distinction of textures from the whole. In principle, both the pairwise and total degree metrics are useful in PATCH analysis. For instance, a community resulting from extensive conservation work could engender a low total external degree due to the difference in tools and artist purpose, while multiple communities from similar sources, such as multiple students of the same master, could show a similar high external degree describing similarity of practice. The internal degree could describe, for instance, the region-to-region consistency of a single artist, or the conformity of practice by multiple similar artist sources. 

\begin{figure}[t]
\includegraphics[width=\textwidth]{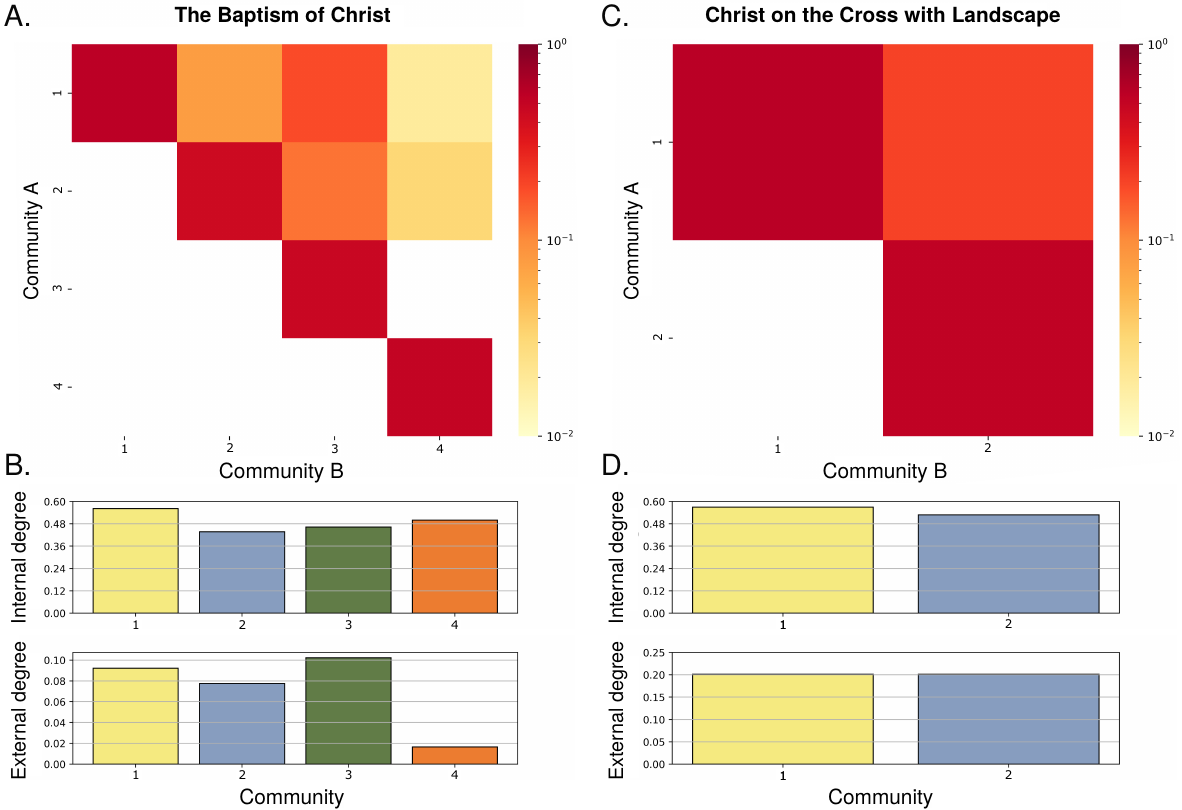}
\caption{Analysis of the network community structure generated
from the PATCH method for: A, B) El Greco's \emph{The Baptism of Christ}
and C, D) \emph{Christ on the Cross with Landscape}. A, C) Normalized network degree between each pair of identified communities. The color scale is logarithmic. B, D) Normalized internal degree (considering all edges within a given community) and total external degree (considering all edges between two distinct communities). The colors of the bars correspond to the communities in the network representations of Figures 3-4. Note that \emph{Baptism} community 4 (orange) in panel B has significantly less external connectivity, and the external degree for the \emph{Christ on the Cross} communities in panel D is significantly higher than for any \emph{Baptism} community.}
\label{fig:degrees}
\end{figure}

Figure \ref{fig:degrees} shows the computed internal and external network degrees for the PATCH analysis of the two \textit{El Greco} paintings. Figure \ref{fig:degrees}~A and C show a pairwise comparison between each of the identified communities in each painting, and Figure \ref{fig:degrees}~B and D show the internal degree and total external degree for each community.

In \emph{The Baptism}, the external degrees between the first three communities are significantly stronger than those involving the fourth community, with the average external degree over four times larger. This indicates a significant isolation of the fourth community (colored orange in Figure 3, and represented by orange bars in Figure \ref{fig:degrees}~B). The strongest external degree is between communities 1 (yellow) and 3 (green) and the weakest is between 3 and 4, which has no connecting edges. Overall, the 3rd community has a slightly greater connection with the remainder of the network, indicating that the height map is the most similar to all other areas. The internal degree is similar for all communities ($< 22$\%), indicating that no community stands out as having a larger disparity in topographical signal. The external degree analysis is largely redundant in the case of \emph{Christ on the Cross}, since only two communities are proposed. However, it should be noted that the external degree between the two communities (\emph{Christ}) is significantly larger than for any other two communities in \emph{The Baptism} except the connection between communities 1 and 3 (\emph{Baptism}). This indicates that, on average, more similarities can be found between the two communities in \emph{Christ on the Cross} than those proposed in \emph{The Baptism}.
This is reflected in the low modularity value. Additionally, the internal degree in each community is similar, which can indicate a similar intra-community stylistic distribution, and supports the two-community model.

\section*{\fontfamily{phv}\selectfont Appendix E: Brush Construction in the Early Modern Period}
In El Greco's time, brushes were made with either bristles
(which are coarse) or hair (which is fine and soft). Each served
distinct technical purposes and often came in material-specific
shapes~\cite{gettens2012painting}. Variation could arise from using the
hair/bristles of different species in the same type of brush. Key
sources were hog bristles (especially white hog) and ``minever'' (gray
squirrel tail fur)~\cite{cennini2018book}
, though dog and mongoose often
replaced minever due to its rarity~\cite{gettens2012painting,palomino1986pictorial}.
Hog's bristles likewise may have been replaced with ox~\cite{fuga2006artists}. 
Such changes could affect
paint application. For example, hog's bristles are naturally flagged
(the ends of the hairs are split), allowing them to pick up more
paint~\cite{hardy1949animal}. 
Bristle substitution could decrease brush
capacity. Further experimentation to understand the effect of materials
on PATCH analysis is needed.

\section*{\fontfamily{phv}\selectfont Appendix F: Additional Figures}

\begin{figure}[h!] \centering \includegraphics[scale=1.65]{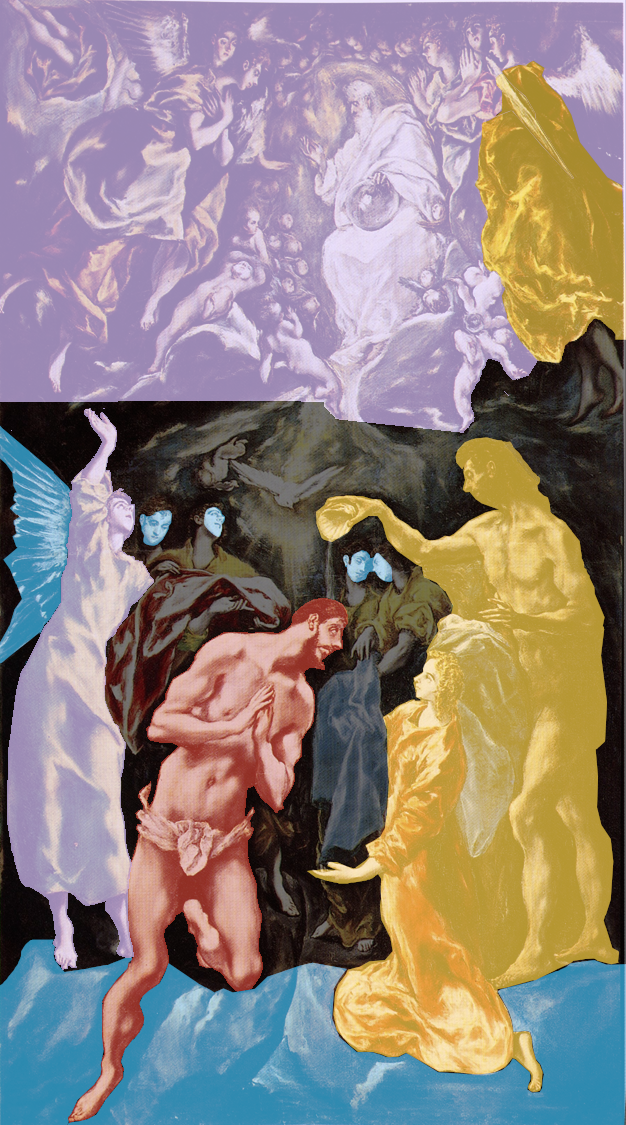}
\caption{Art historical attributions of portions of the painting. Areas attributed to El Greco are overlaid in purple, Jorge Manuel in gold, both artists in red, and a third hand or hands in light blue. }
\end{figure}

\begin{figure}[h!] \centering \includegraphics[width=0.8\textwidth]{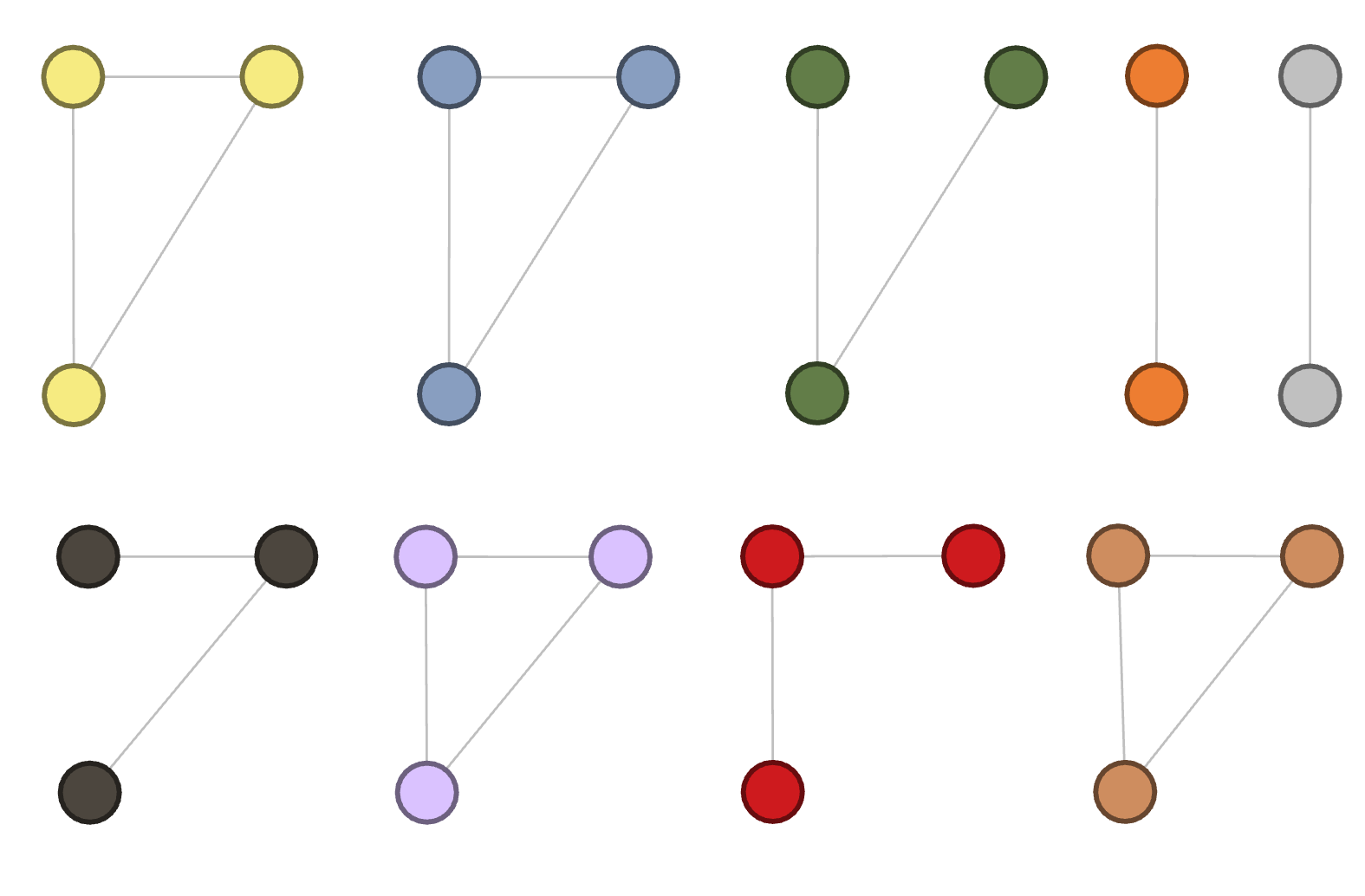}
\caption{Network diagram for the student paintings with spurious edges trimmed. Nine disjoint communities are identified, corresponding to the individual artists.}
\end{figure}

\section*{\fontfamily{phv}\selectfont Appendix G: Edge Pruning}
The PATCH method is concerned with ``sameness,'' meaning we need a way
to remove spurious edges from the network. To accomplish this, we employ
an edge pruning process based on uniqueness. We assume that regions with
a high degree (number of connections) have more common features and
nodes with a low degree have more unique features. Therefore, a
connection between regions with more unique features is more likely to
be correct. We operationalize this assumption by assigning each edge a
score based on the average uniqueness of its nodes, where uniqueness is
defined as the percentage of nodes a node does not connect with:
$W_{ij} = \frac{1}{2}((\text{max} - \text{degree}_{i})/\text{max} + (\text{max} -
\text{degree}_{j})/\text{max})$ where max = the maximum possible degree
(i.e., number of connections) and degree = the degree
of one of the two nodes in question. There were two erroneous
same-artist pairs in our experimental dataset (9\% of edges in the
network), and these edges had the lowest uniqueness scores. Thus, we
trim the 9\% of edges (or more if there are ties) in a given network
with the lowest uniqueness scores.


\end{document}